\documentclass[cameraready]{vgtc}

\ifpdf
  \pdfoutput=1\relax                   
  \usepackage{graphicx}                
  \DeclareGraphicsExtensions{.pdf,.png,.jpg,.jpeg} 
\else
  \usepackage{graphicx}                
  \DeclareGraphicsExtensions{.eps}     
\fi%

\graphicspath{{figures/}{pictures/}{images/}{./}} 

\usepackage{microtype}                 
\PassOptionsToPackage{warn}{textcomp}  
\usepackage{textcomp}                  
\usepackage{mathptmx}                  
\usepackage{times}                     
\usepackage{cite}                      
\usepackage{tabu}                      
\usepackage{booktabs}                  
\usepackage{subfig}
\usepackage{svg}
\renewcommand\footnotemark{}

\usepackage{multirow}

\usepackage{float}

\usepackage{tabularx}

\usepackage{colortbl}
\usepackage{xcolor}

\onlineid{1307}

\vgtccategory{Research}

\vgtcinsertpkg

\usepackage{amsmath}
\usepackage{hyperref}
\usepackage{cleveref} 
\usepackage{mdframed}
\usepackage{makecell}
\usepackage{url}
\usepackage{enumitem}
\usepackage{float}

\title{Advancing the Understanding and Evaluation of AR-Generated Scenes: \\ When Vision-Language Models Shine and Stumble}

\author{Lin Duan$^{1*}$%
\and Yanming Xiu$^{2*}$%
\and Maria Gorlatova$^{3}$}

\affiliation{\scriptsize Department of Electrical and Computer Engineering, Duke University\thanks{\parbox{\textwidth}{\noindent *Equal contribution. \\ \{$\rm{lin.duan}^1, \rm{yanming.xiu}^2, \rm{maria.gorlatova}^3$\}@duke.edu}}}

\abstract{
Augmented Reality (AR) enhances the real world by integrating virtual content, yet ensuring the quality, usability, and safety of AR experiences presents significant challenges. Could Vision-Language Models (VLMs) offer a solution for the automated evaluation of AR-generated scenes? In this study, we evaluate the capabilities of three state-of-the-art commercial VLMs---GPT, Gemini, and Claude---in identifying and describing AR scenes. For this purpose, we use DiverseAR, the first AR dataset specifically designed to assess VLMs' ability to analyze virtual content across a wide range of AR scene complexities. Our findings demonstrate that VLMs are generally capable of perceiving and describing AR scenes, achieving a True Positive Rate (TPR) of up to 93\% for perception and 71\% for description. While they excel at identifying obvious virtual objects, such as a glowing apple, they struggle when faced with seamlessly integrated content, such as a virtual pot with realistic shadows. Our results highlight both the strengths and the limitations of VLMs in understanding AR scenarios. We identify key factors affecting VLM performance, including virtual content placement, rendering quality, and physical plausibility. This study underscores the potential of VLMs as tools for evaluating the quality of AR experiences. 

} 

\CCScatlist{
    \CCScatTwelve{Mixed / Augmented Reality}{Vision Language Models}{Content Evaluation}{Scene Understanding};
}

\begin{document}



\firstsection{Introduction}

\maketitle


Augmented Reality (AR) is poised to transform how we interact with the world, unlocking groundbreaking applications across education, entertainment, and healthcare. However, some AR applications can negatively impact user experience, either unintentionally or through deliberate malice. For example, AR content that does not blend seamlessly with the physical environment can reduce immersion and undermine the intended objectives of the AR experience, due to issues like spatial misalignment \cite{wasenmuller2016augmented} or stylistic mismatches \cite{style02}. Additionally, some AR content may mislead or confuse users \cite{misleading01}. These challenges not only compromise user experience but also raise safety and ethical concerns. Therefore, evaluating the quality of AR content has become a critical challenge to ensure usability and safety in AR applications.

While prior studies have investigated methods for assessing general image quality~\cite{lin2018hallucinated, wang2023exploring}, evaluating AR scenes introduces unique challenges that surpass those of traditional image analysis. Unlike conventional approaches that primarily rely on mathematical metrics (e.g., SSIM~\cite{hore2010image}) or, more recently, machine learning models (e.g., IQAGPT~\cite{chen2023iqagpt}), the quality of an AR scene is deeply intertwined with the user's subjective experience. This is affected by factors such as content placement, lighting, rendering quality, and the seamless integration of virtual elements into the physical environment.  Typically, these aspects are assessed through user studies~\cite{merino2020evaluating}, which gather feedback after users interact with the AR app. However, this approach has some drawbacks: experiencing an AR scene is often time-consuming, and variations in real-world environments or AR content during interactions can skew the evaluation. Real-time AR scene quality estimation offers a promising alternative, enabling immediate and holistic assessments. Yet, collecting real-time user feedback may not always be practical, as it could disrupt the interaction and detract from the experience. This highlights the need for objective and real-time methods to automatically evaluate AR content.

\begin{figure}[t]
\centering
\vspace{0.2cm}
\includegraphics[width=0.47\textwidth ]{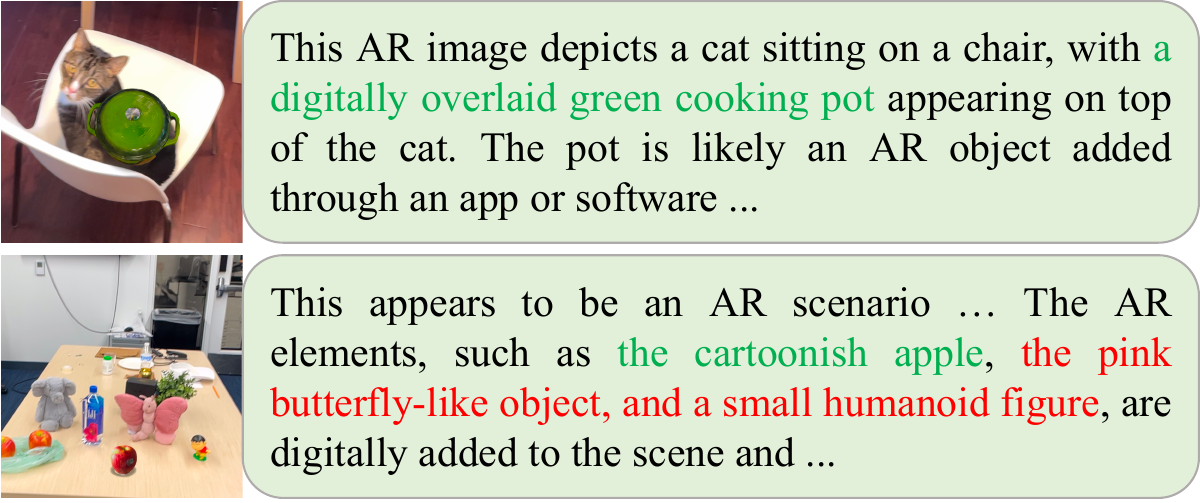}
\vspace{-0.3cm}
\caption{AR content is integrated into real-world contexts: a virtual pot placed on a cat (top) and a virtual apple placed on a table (bottom). GPT can accurately recognize and describe the AR pot on the cat but mistakenly identifies the real toys as AR elements.}
\label{figure:vlmability}
\vspace{-0.6cm}
\end{figure}

Recent developments in Vision Language Models (VLMs) present potential solutions to the challenges of AR experience evaluation. These models enable a holistic, human-like understanding of complex scenes~\cite{vlmreview}, excelling in context-aware analysis by capturing relationships between objects and generating detailed descriptions that align with human perception. 
However, it is essential to understand VLMs' ability toward AR content understanding before using them for AR content evaluation. 
Examples in Figure~\ref{figure:vlmability} present AR scenes together with GPT's responses to the question, ``Is this an augmented reality (AR) scene? If yes, what virtual content does it include, and could you explain why each element is considered virtual?" These examples suggest that VLMs can recognize AR images but are sometimes unreliable. 
To address this, in this work, we provide an 
assessment of the feasibility of using modern VLMs for evaluating AR content. Toward this goal, we collected AR images from different sources and tested three state-of-the-art VLMs, GPT~\cite{GPT4V}, Gemini~\cite{gemini}, and Claude~\cite{anthropic2024claude3}, with different prompt questions on these images. The contributions of this paper include:

\vspace{-0.2cm}
\begin{itemize}[noitemsep] 

    \item We curate and release a dataset\footnote{\url{https://github.com/ARResearch-1/DiverseAR-Dataset}\label{footnote:dataset}}, DiverseAR, comprising 318 images collected from one public website~\cite{DeepAR}, two commercial AR platforms (Amazon and Scaniverse), three AR applications previously developed by our lab running on Magic Leap, Android, and HoloLens~\cite{qu2024looking, xiu2025viddar, eom2022neurolens}, along with two custom-built AR applications for this project, running on Apple Vision Pro and Android. To the best of our knowledge, this dataset is \textbf{the first of its kind for assessing the capabilities of VLMs in identifying and describing AR content}, encompassing diverse AR scenarios.
    
    \item We test VLMs on identifying and describing AR content. Results show that VLMs are effective in recognizing and describing obvious and typical AR content, reaching a perception TPR of up to 93\% and description TPR of up to 71\% across varying AR scene complexity levels within the DiverseAR dataset. 
    
    \item We observe that VLMs face challenges when encountering well-designed and seamlessly integrated AR content. As the AR scene complexity of images in DiverseAR increases, VLMs' perception and description TPRs experience a noticeable decline, dropping from 75.8\% to 22.8\% and from 97.8\% to 34.2\%, respectively.
    

    \item We conducted a user study with IRB approval to compare human and VLM performance in identifying and describing AR scenarios, uncovering differences and similarities in AR experience evaluation and reasoning patterns.

\end{itemize}
\vspace{-0.3cm}

Below, we review related work in~\Cref{sec:related work}, introduce the DiverseAR dataset in~\Cref{sec:dataset}, describe our methods in~\Cref{sec:method}, evaluate VLMs in~\Cref{sec:results}, and conclude the paper in~\Cref{sec:discussion}. 

\begin{table*}[t]
\centering
\caption{Key characteristics of the virtual and real objects in the DiverseAR dataset.}
\vspace{-0.2cm}
\begin{tabular}{|l|l|} 
\hline
\textbf{Characteristic} & \textbf{Description}  \\ 
\hline
Placement variations & Real and virtual objects positioned in typical (e.g. apple on a table) and abnormal (e.g. pot on a cat) places.  \\ 
\hline
Physical law violations & Floating objects without realistic grounding and intersecting objects where real and virtual objects overlap. \\
\hline
Size and shape variations & Virtual objects exhibit diverse geometries, including varying sizes and shapes. \\
\hline
Shadow variations &  Shadows range from realistic (accurate rendering) to flawed and completely absent. \\
\hline
Lighting variations & Lighting setups include a range from bright to dim and scenarios with glowing or non-glowing objects. \\
\hline
Transparency variations & Virtual objects with varying levels of transparency. \\
\hline
Render quality variations & Virtual content with low and high visual fidelity. \\
\hline
\end{tabular}
\label{table:diverseARdataset}
\vspace{-0.5cm}
\end{table*}
\section{Related Work}
\label{sec:related work}

Recently, the advancement of VLMs has made them a transformative technology, capable of providing comprehensive analysis of complex scenes. This positions VLMs as efficient tools for evaluating AR content quality. Previous studies have explored their utility in various domains, including analyzing image perception for advertisements and medical images~\cite{malakouti2024benchmarking, chen2023iqagpt}, detecting visual anomalies in the generated images~\cite{zhou2023rome, bitton2023breaking}, and assessing the look and feel of images~\cite{wang2023exploring}, highlighting their potential for content assessment. Yet, to our knowledge, no prior efforts have applied VLMs to evaluate AR content. 

Despite their promise, concerns persist regarding VLMs' perception and reasoning capabilities. Studies have shown that VLMs can hallucinate, producing information not grounded in the data~\cite{hallucination01}. Additionally, VLMs exhibit limited ability to comprehend depth information~\cite{cai2024spatialbot}, further raising questions about their reliability in evaluating the quality of AR experiences, particularly in aspects such as spatial alignment and contextual integration. In light of these issues, this work evaluates the ability of VLMs to analyze and interpret AR images, offering a pioneering investigation into their effectiveness in AR content evaluation.

\section{DiverseAR Dataset}
\label{sec:dataset}

In this section, we introduce \textbf{DiverseAR}, a new dataset curated as part of this work. The dataset is publicly available on GitHub\textsuperscript{\ref{footnote:dataset}}.

\subsection{Motivation and Collection Process}

To evaluate the AR scene understanding capabilities of VLMs, we curated the DiverseAR dataset, specifically designed to capture a broad spectrum of AR scenarios. The dataset consists of \emph{298 AR images} collected from diverse sources and environments. It includes 23 images captured using a custom-developed Apple Vision Pro AR application in laboratory and kitchen environments, and 151 images collected from a custom-developed Android AR application in bedroom and dining room environments. Additionally, 42 images were collected during the exploration of AR-specific research topics, such as attention patterns\cite{qu2024looking}, virtual content arrangements\cite{xiu2025viddar}, and surgical guidance\cite{eom2022neurolens}. The dataset also features 7 images from the Amazon app's AR view and 46 images collected from the Scaniverse\cite{niantic2024scaniverse} app's AR view, captured in laboratory, kitchen, and dining room environments. Finally, 29 images were sourced from a website showcasing AR advertisement videos~\cite{DeepAR}. Additionally, we included \emph{20 non-AR images} to supplement the dataset.

\subsection{Dataset Composition}

\begin{figure}[t]
\centering
\vspace{0.2cm}
\includegraphics[width=0.47\textwidth ]{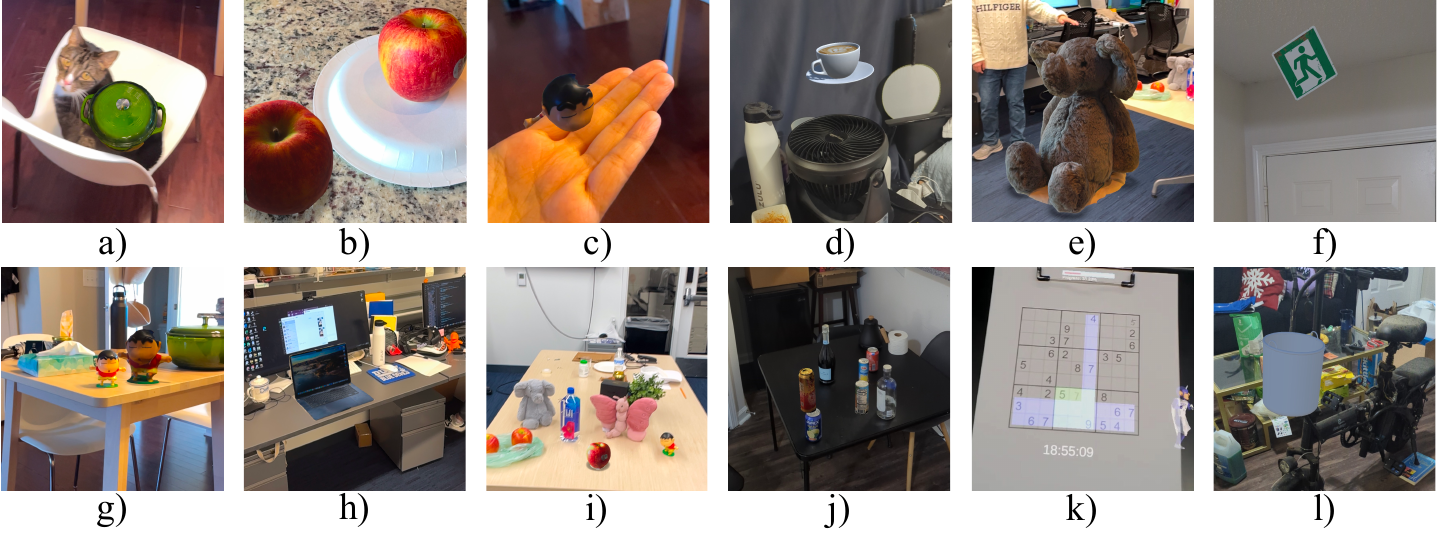} 
\vspace{-0.5cm}


\caption{Examples of key characteristics of the virtual and real objects in the DiverseAR dataset. 
a) Virtual pot positioned in an abnormal place (on a cat);
b) Real and virtual apples positioned in a typical place (table);
c) Virtual toy intersecting a real palm;
d) Floating virtual cup;
e) Virtual toy with unusual size;
f) Virtual sign in an abnormal pose;
g) Large virtual toy on the table with its shadow cast in the wrong direction;
h) Virtual laptop with no shadow;
i) Virtual apple alongside real objects under bright lighting;
j) Three virtual cans alongside real objects under dim lighting;
k) Transparent, glowing virtual highlights;
l) Low-render-quality virtual basket.}
\label{figure:diverseARdataset}
\vspace{-0.6cm}
\end{figure}

The DiverseAR dataset encompasses a wide spectrum of AR and non-AR scenarios, showcasing diverse characteristics of both virtual and real-world content. Table~\ref{table:diverseARdataset} summarizes the key characteristics represented in the dataset, while Figure~\ref{figure:diverseARdataset} illustrates representative examples of these characteristics. The dataset includes single or multiple instances of identical and varied real and virtual objects across various AR images. These objects span various classes, such as toys, food, shoes, plants, laptops and containers.

By offering this diversity in both content and context, the DiverseAR dataset enables an insightful evaluation of VLMs in interpreting simple and complex AR scenes. 

\section{Methods}
\label{sec:method}

To evaluate the AR scene understanding capabilities of VLMs and identify factors influencing their performance, we assess three commercial VLMs using our DiverseAR dataset: 1) \textbf{GPT-4o-2024-08-06} by OpenAI; 2) \textbf{Claude-3.5-Sonnet-20241022} by Anthropic; 3) \textbf{Gemini-1.5-Pro-002} by Google. These models represent state-of-the-art capabilities in vision-language understanding and generation. We also design targeted prompts to assess the accuracy of their perception and description of AR scenes. Additionally, we define three levels of AR scene complexity based on AR content attributes and categorize AR samples into five groups based on VLM responses. These categories facilitate a detailed quantitative analysis of VLM performance in recognizing and describing AR scenes. 




\subsection{Prompt Design for AR Scene Understanding}
\label{subsec:Prompt Design for AR Scene Understanding}

To assess the capacity of VLMs to generate comprehensive textual descriptions of images, we design two types of prompts: \textbf{general} and \textbf{task-aware}.

\noindent\textbf{General Image Captioning Prompt:} ``\textit{Can you explain what is happening in this image?}" 
    
This prompt is designed to assess the model's ability to generate a general caption of the image without any specific guidance toward identifying AR content. It tests whether the model can naturally recognize and describe AR elements without explicit textual cues.
    
\noindent\textbf{Task-aware Image Captioning Prompts:} 1) \textbf{Without explicitly referencing AR:} ``\textit{Does this image include virtual content superimposed on the real world? If yes, could you list all the virtual elements present and explain why each one is considered virtual?}" 2) \textbf{Explicitly referencing AR:} ``\textit{Is this an augmented reality (AR) scene? If yes, what virtual content does it include, and could you explain why each element is considered virtual?}"
    
These task-aware image captioning prompts are designed to assess the model's ability to leverage guidance for more accurate identification and description of AR-specific elements. Since the responses from VLMs were similar for both task-aware prompts, we chose to use the prompt without explicitly referencing AR in our experiments for consistency and simplicity.

\subsection{Classification of AR Scene Complexity Levels}
\label{subsec:Classification of AR Scene Difficulty Levels}


To analyze VLM performance across varying complexities of AR scenes, we define three distinct levels based on the attributes of virtual elements as follows:

\vspace{-0.3cm}
\begin{itemize}[noitemsep]
    \item \textbf{Easy:} Images with obvious virtual content, such as transparent or glowing overlays, or virtual objects with low rendering quality that are easily distinguishable from the real world.
    \item \textbf{Medium:} Images with high-quality virtual content that exhibits inconsistencies with physical laws, such as floating or intersecting objects, or virtual objects with unrealistic attributes like informal size or placement relative to the real world.
    \item \textbf{Hard:} Images with high-quality virtual content seamlessly integrated into the real-world environment, including proper shadows, realistic size and shape, and adherence to physical laws, making them more challenging to distinguish as virtual.
\end{itemize}
\vspace{-0.3cm}

Labeled by an annotator with extensive expertise in AR, there are 91, 128, and 79 images in the easy, medium, and hard levels, respectively. 


\subsection{Categorization of VLM Responses}
\label{subsec:Categorization of VLM Responses}
To facilitate the analysis of factors affecting VLM performance, we categorize AR samples into five groups based on the VLMs' responses:
\vspace{-0.2cm}
\begin{itemize}[noitemsep]
    \item \textbf{Category 1: Accurate AR Recognition.} AR images are correctly identified as AR scenes, and the virtual content is accurately recognized and described. This includes proper classification of real versus virtual elements, their attributes, and their spatial relationships.
    \item \textbf{Category 2: Partial AR Recognition.} AR images are recognized as AR scenes, but errors in identifying virtual content persist, including misclassifying real elements as virtual (or vice versa), incorrect object classification, or inaccurate descriptions of object properties such as location.
    \item \textbf{Category 3: Missed AR Recognition.} AR images are not recognized as AR scenes. Instead, they are misidentified as real-world images or as digital content displayed on screens, failing to acknowledge the presence of AR elements.
    \item \textbf{Category 4: False AR Detection.} Non-AR images are incorrectly identified as AR scenes. The VLM mistakenly attributes virtual content to images that contain only real-world elements.
    \item \textbf{Category 5: Accurate Non-AR Recognition.} Non-AR images are correctly identified as not being AR scenes, with no virtual content mistakenly attributed to them.
\end{itemize}
\vspace{-0.3cm}

We apply this categorization across the whole dataset. For each group, we analyze the key characteristics of the samples that influence VLM performance in~\Cref{sec:results}.

\subsection{Evaluation Metrics}
\label{subsec:Performance of VLMs}

\noindent\textbf{Metrics for AR Scene Perception:}
To assess whether VLMs can accurately identify AR scenes, we evaluate their performance using the True Positive Rate, \(TPR_P\),  and the True Negative Rate, \(TNR_P\). These metrics are formulated as follows:
\vspace{-0.2cm}

\[
TPR_P = \frac{N_1 + N_2}{N_1 + N_2 + N_3}, TNR_P = \frac{N_5}{N_4 + N_5},
\]
\vspace{-0.2cm}

\noindent where \(N_1\), \(N_2\), \(N_3\), \(N_4\), and \(N_5\) denote the number of images in Category 1, 2, 3, 4, and 5, respectively. \(TPR_P\) represents the proportion of AR images identified as containing AR elements among all AR images, regardless of the accuracy of the AR scene descriptions. Similarly, \(TNR_P\)  indicates the proportion of non-AR images described without referencing AR elements among all non-AR images, irrespective of the accuracy of the non-AR scene descriptions.

We use these metrics instead of simple accuracy because the DiverseAR dataset contains relatively few non-AR images, which were collected only to supplement the study. These metrics enable a more nuanced evaluation, especially for imbalanced datasets. 

\noindent\textbf{Metric for AR Scene Description:}
To evaluate AR scene description performance, we measure the VLMs' ability to correctly identify and describe virtual content within AR scenes. This is assessed using the metric \(TPR_D\), which is defined as follows:
\vspace{-0.2cm}
\[
TPR_D = \frac{N_1}{N_1 + N_2 + N_3},
\]
\vspace{-0.3cm}

\noindent\(TPR_D\) represents the percentage of AR images that are both correctly identified as containing AR elements and accurately described in terms of the AR scene, out of all AR images. This metric, alongside perception metrics, allows for a detailed evaluation of the VLMs' strengths and limitations in AR scene understanding. By focusing on the quality of both perception and description, we aim to provide detailed insights into VLM performance across varying levels of AR scene complexity. These results are analyzed further in~\Cref{sec:results}.

\vspace{-0.1cm}

\section{Results}
\label{sec:results}

\begin{figure*}[t]
\centering
\vspace{0.2cm}
\includegraphics[width=0.98\textwidth ]{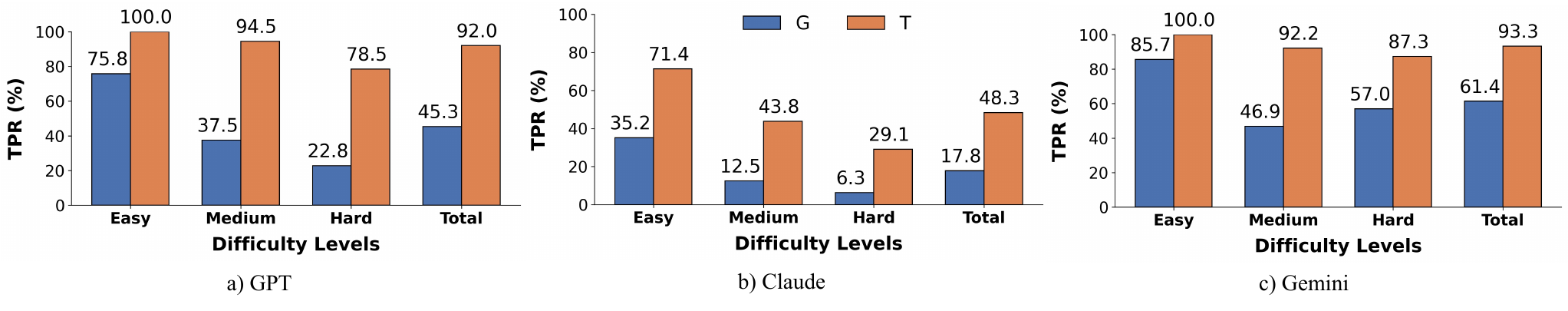}
\vspace{-0.5cm}
\caption{The True Positive Rate for Perception (\(TPR_P\)) of the three VLMs across varying complexity levels using G (general) and T (task-aware) image captioning prompts. VLMs show a consistent decline in performance as AR scene complexity increases.}
\label{fig:perception_results_level}
\vspace{-0.5cm}
\end{figure*}

\begin{figure*}[t]
\centering
\vspace{0.2cm}
\includegraphics[width=0.98\textwidth ]{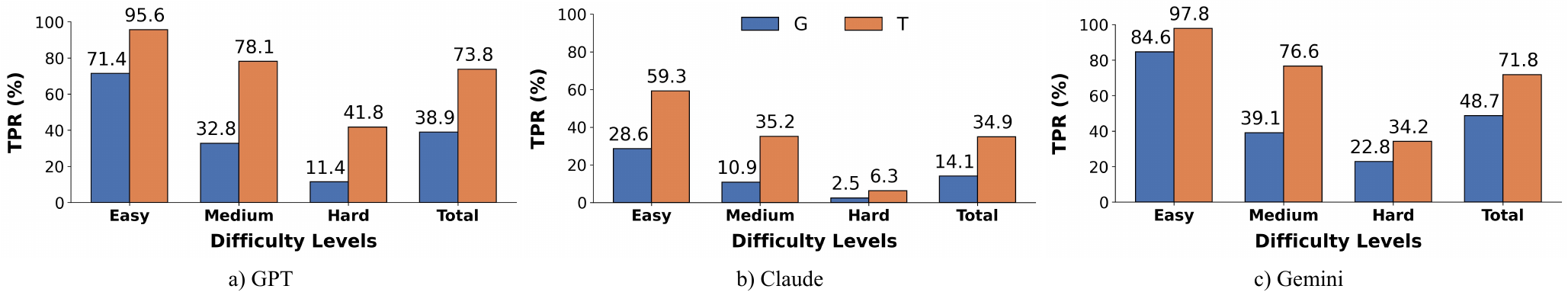}
\vspace{-0.5cm}
\caption{The True Positive Rate for Description (\(TPR_D\)) of the three VLMs across varying AR scene complexity levels using G (general) and T (task-aware) image captioning prompts. Notably, GPT outperforms both Gemini and Claude on medium and hard AR scenes using prompt T.}
\label{fig:description_results_level}
\vspace{-0.6cm}
\end{figure*}

We present the results of VLMs' AR scene understanding capabilities, focusing on their perception and description performance. We also provide an in-depth categorical analysis of VLM responses and compare their performance with that of human counterparts.

\subsection{Performance of VLMs' AR Understanding Abilities}
\noindent \textbf{Perception Performance:} 

We evaluate VLMs' ability to perceive AR scenes at different complexity levels, using the DiverseAR dataset as well as the prompts defined in~\Cref{sec:method}. 

Across the dataset, the Perception True Negative Rate (\(TNR_P\)) remains consistently at 100\% for all models. This demonstrates that VLMs rarely misclassify real-world images as AR content, showcasing their robustness in distinguishing non-AR scenes.

When comparing the Perception True Positive Rate (\(TPR_P\)) using the task-aware prompt (T) and the general prompt (G), we observe significant improvements with prompt T across all models, as illustrated in Figure~\ref{fig:perception_results_level}. We observe performance improvements ranging from 30.5\% to 46.7\%, \emph{highlighting the importance of explicitly specifying AR-related textual cues}. Among the three models we evaluated, Gemini consistently achieves the highest \(TPR_P\), with 61.4\% under prompt G and 93.3\% under prompt T, outperforming GPT and Claude. These results underscore Gemini’s superior performance and its potential suitability for AR scene perception tasks.

Further analysis of \(TPR_P\) across different AR scene complexity levels reveals a consistent decline in performance as complexity increases. For instance, under prompt G, GPT’s \(TPR_P\) decreases from 71.4\% for easy scenes to just 11.4\% for hard scenes, with similar trends observed for all models. This suggests that while VLMs excel at identifying simpler AR objects (e.g., easy-level AR scenes), their performance deteriorates when faced with more complex AR content characterized by polished rendering and realistic integration.

\noindent \textbf{Description Performance:} 
In addition to perception, we evaluate the correctness of VLMs' virtual content descriptions across varying scene complexity levels and the full dataset. 

Similar to perception, the Description True Negative Rate (\(TNR_D\)) remains consistently at 100\% across all models and the dataset. This indicates that VLMs do not mistakenly describe non-AR scenes as AR content, even in scenarios involving unusual object placements.

The prompt T significantly improves description accuracy compared to the prompt G, as shown in Figure~\ref{fig:description_results_level}. For instance, GPT's Description True Positive Rate (\(TPR_D\)) increases from 38.9\% with prompt G to 73.8\% with prompt T, emphasizing the importance of tailored prompts. While Gemini outperforms GPT and Claude under prompt G, GPT achieves the highest \(TPR_D\) under prompt T (73.8\%), surpassing Gemini (71.8\%) by 2\%. This result indicates that \emph{GPT has the potential to leverage explicit task prompts more effectively when analyzing AR scenes}. In contrast, Claude consistently underperforms, with \(TPR_D\) as low as 14.1\% under prompt G, improving to only 34.9\% under prompt T. Detailed analysis reveals that Claude often generates high-level summaries of objects in the scene without capturing the relationships among them, limiting its ability to accurately describe AR-specific content.

Analyzing \(TPR_D\) across different levels of complexity reveals a consistent decline in performance as complexity increases, mirroring the trends observed in perception results. Notably, GPT outperforms both Gemini and Claude on medium and hard AR scenes. This highlights GPT’s strength in tasks requiring detailed AR content descriptions, particularly in more complex scenes where spatial reasoning and contextual understanding are critical.



\subsection{Categorical Analysis of VLMs' Responses}
We analyze AR samples across the five response categories defined in~\Cref{sec:method}, examining key observations, contributing factors, and potential reasons for the VLMs' behavior.

\noindent\textbf{Category 1: Accurate AR Recognition.} In this category, VLMs demonstrate strong performance in identifying AR objects. We observe that AR content with characteristics such as text overlays, low-quality rendering (e.g., uniform pixel distributions), or implausible physical attributes (e.g., lighting, shape, size mismatches, unnatural object placement like floating objects, or unrealistic transparency) is often recognized correctly, as shown in Figure~\ref{figure:diverseARdataset}(k, l, e, d). These attributes are prevalent in common AR scenarios and are likely well-represented in the VLMs' training datasets, making the models adept at detecting them.

\begin{figure}[t]
\centering
\vspace{0.2cm}
\includegraphics[width=0.48\textwidth ]{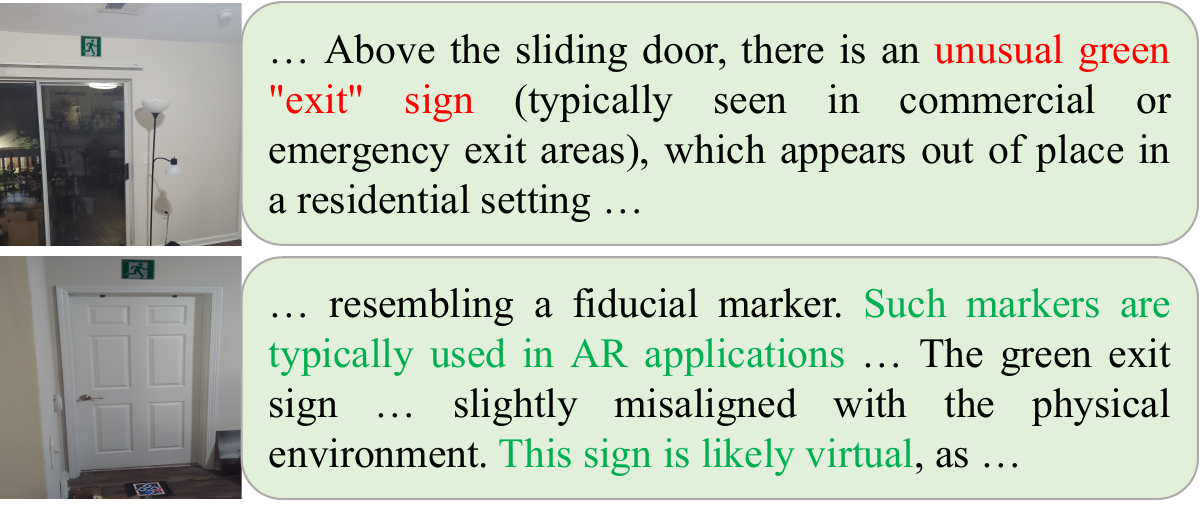}
\vspace{-0.6cm}
\caption{These images feature an AR-generated ``Exit" sign integrated into real-world contexts.  With the prompt T, GPT accurately identifies the sign with a QR code (bottom) but misclassifies it as a real object without the code (top).}
\label{fig:sign}
\vspace{-0.6cm}
\end{figure}

\begin{figure}[t]
\centering
\vspace{0.2cm}
\includegraphics[width=0.48\textwidth ]{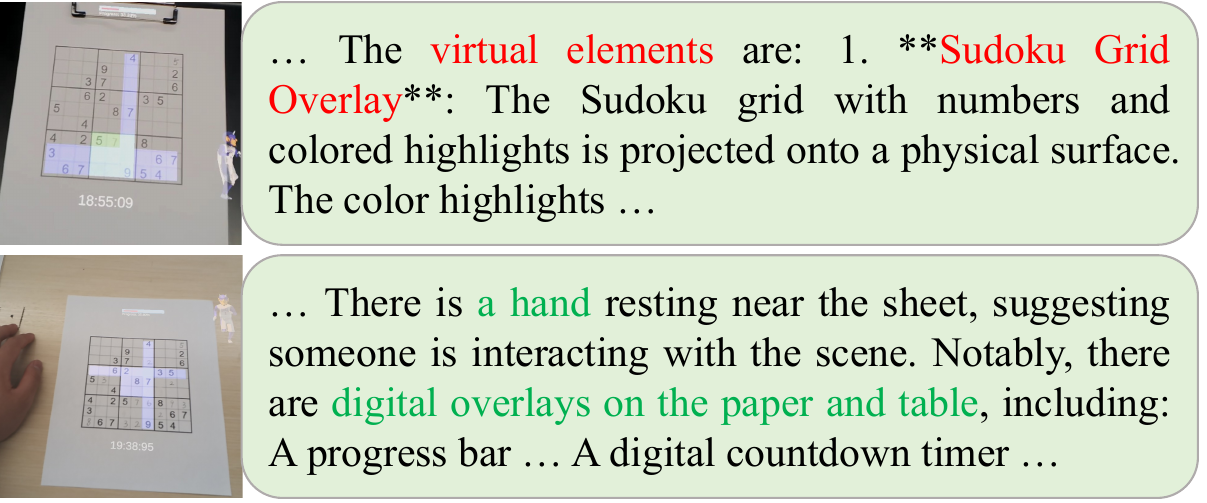}
\vspace{-0.6cm}
\caption{These images feature AR-generated highlights, avatars, and time and completion bars integrated into a real-world Sudoku puzzle. With the prompt T, GPT accurately recognizes AR content when hands are present (bottom) but misidentifies the real grid as virtual content in their absence (top). }
\label{fig:sudoku}
\vspace{-0.6cm}
\end{figure}

Interestingly, \emph{certain visual cues enhance recognition accuracy}. For instance, the presence of QR codes or hands provides strong contextual clues. As shown in Figure~\ref{fig:sign}, a virtual sign with a QR code is correctly identified, while the same virtual sign without it is misclassified. Similarly, Figure~\ref{fig:sudoku} shows that the presence of hands improves AR recognition, while their absence causes VLMs to misclassify real grids as virtual content. Additionally, implausible physical interactions, such as a pot placed on a cat, further help the models identify AR content. The ability of VLMs to identify these features can be attributed to the self-attention mechanism in transformer-based architectures~\cite{han2022survey}, which processes an image as a sequence of patches, capturing global context and spatial relationships. This design enables the models to effectively understand and contextualize relationships, actions, and attributes within an image.

\begin{figure}[t]
\centering
\vspace{0.2cm}
\includegraphics[width=0.48\textwidth ]{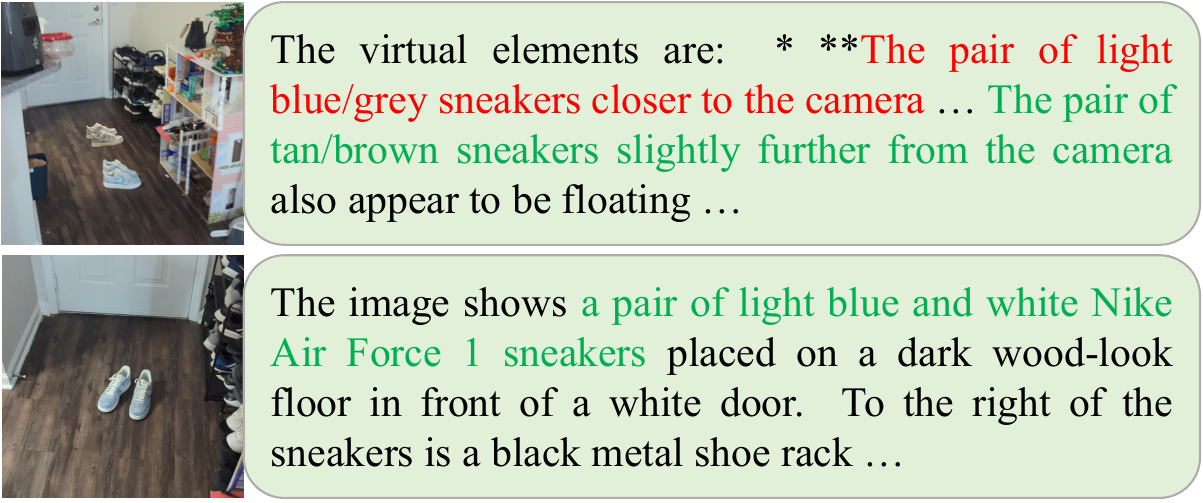}
\vspace{-0.6cm}
\caption{These images showcase an AR-generated image with a virtual pair of sneakers placed next to real sneakers (top) and a non-AR image displaying only real sneakers (bottom). With prompt T, Gemini correctly identifies a real pair of sneakers in the absence of virtual sneakers (bottom) but misclassifies them as virtual when virtual sneakers are added to the scene (top).}
\label{fig:shoe}
\vspace{-0.6cm}
\end{figure}

\begin{figure}[t]
\centering
\vspace{0.2cm}
\includegraphics[width=0.48\textwidth ]{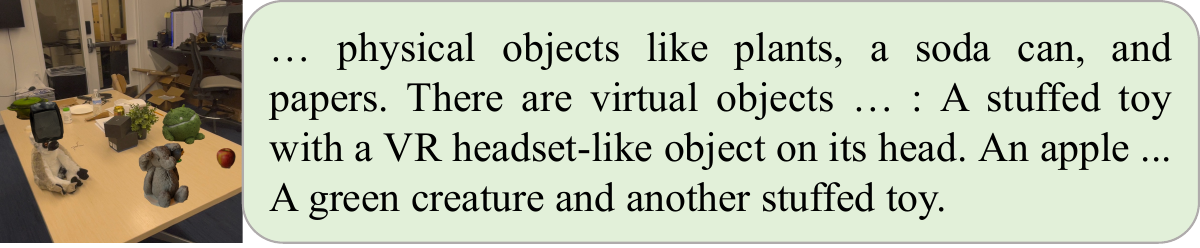}
\vspace{-0.6cm}
\caption{This image features an AR-generated pot, heater, apple, toy elephant, and toy android integrated into a real-world context. With prompt
T, GPT fails to identify a virtual toy elephant and a virtual pot with realistic shadows as virtual objects.}
\label{fig:mix}
\vspace{-0.6cm}
\end{figure}

\noindent\textbf{Category 2: Partial AR Recognition.} In this category, VLMs misinterpret AR content, in three primary ways. First, real objects are recognized as virtual. For instance, as shown in Figure~\ref{fig:shoe}, a real pair of sneakers is identified correctly when no virtual sneakers are present. However, adding virtual sneakers to the scene causes the VLM to misclassify the real sneakers as virtual. Second, virtual content is only partially recognized. For example, in a mixed scene, a virtual toy elephant and a virtual pot are not recognized as virtual objects, as illustrated in Figure~\ref{fig:mix}. Shadows rendered on these virtual objects can enhance their realism, allowing them to evade detection. Lastly, virtual objects may be misclassified entirely or have their properties misunderstood. For example, a toy butterfly might be incorrectly identified as another type of object, possibly due to gaps in the training data. These issues are most common when virtual and real elements are combined, creating ambiguity that misleads VLMs. Consequently, the model's response to a certain object can vary depending on its surrounding elements, suggesting that these errors likely stem from the spatial reasoning capabilities of transformer-based models. 

\begin{figure}[t]
\centering
\vspace{0.2cm}
\includegraphics[width=0.45\textwidth ]{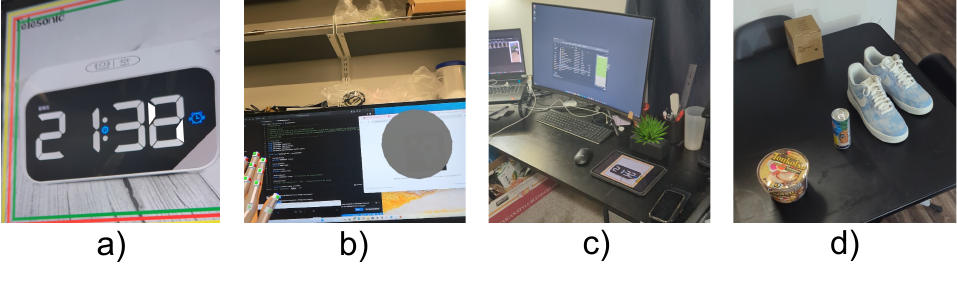}
\vspace{-0.5cm}
\caption{Representative examples of Category 3 (Missed AR Recognition) and Category 5 (Accurate Non-AR Recognition). 
a) Virtual white bars placed on digital screens (Category 3);
b) A virtual circle with a low transparency level in front of the screen (Category 3);
c) A virtual plant is recognized as a real-world fake plant (Category 3);
d) A pair of real sneakers is placed next to the food on the table (Category 5).}
\label{figure:cate3}
\vspace{-0.6cm}
\end{figure}

\noindent\textbf{Category 3: Missed AR Recognition.}
Failures in this category arise when realistic virtual content adheres to physical laws or is seamlessly integrated into real-world settings. These scenarios can mislead VLMs, making it difficult for them to identify virtual elements. For instance, virtual objects with realistic shadows frequently evade detection, especially in environments with weak visual cues. These include dark surfaces, low-light conditions, or inverse lighting directions (Figure~\ref{figure:diverseARdataset}(j)). Similarly, virtual objects positioned on digital screens can also mislead VLMs, as seen in Figure~\ref{figure:cate3}(a). 

We also observe several interesting cases within this category. For example, virtual objects with inadequate transparency are misidentified as real (Figure~\ref{figure:cate3}(b)). Some virtual content with low rendering quality may be misclassified as realistic ``fake objects", as such ``fake objects" are often encountered in daily life (Figure~\ref{figure:cate3}(c)). Likewise,  objects that defy physical laws—such as floating slightly but remaining attached to the surfaces of real objects—still appear plausible enough to evade detection (Figure~\ref{figure:diverseARdataset}(j)). Moreover, virtual objects that do not typically appear in AR settings further exacerbate the recognition problem (Figure~\ref{figure:diverseARdataset}(h)).

These observations highlight several limitations of VLMs. For instance, they struggle with depth perception in 2D images and have trouble interpreting situations that challenge common-sense knowledge~\cite{zhou2023rome}, making it difficult to identify violations of some physical laws. Additionally, we speculate that many models may have limited exposure to diverse or atypical AR-specific samples during training, potentially hindering their ability to effectively understand AR scenes.

\noindent\textbf{Category 4: False AR Detection.} No instances of false AR detection were observed, showcasing the model's robustness in accurately distinguishing non-AR images. To further investigate the generalizability of this observation, our ongoing work involves constructing real-world scenarios with implausible lighting and object placements. Examples include shining a light onto a real apple or attaching a pen unnaturally to a shelf.

\noindent\textbf{Category 5: Accurate Non-AR Recognition.}  
VLMs show strong performance in recognizing non-AR images. Notably, even real-world objects with unconventional placements (Figure~\ref{figure:cate3}(d)) were correctly identified as real rather than virtual, suggesting that VLMs are less likely to be confused by physically plausible yet atypical arrangements.



\subsection{User Study}
To compare human performance with VLMs in identifying and describing AR scenarios, we conducted a user study approved by the Duke University Campus Institutional Review Board (protocol number: 2020-0292). We recruited five participants from the Duke community, comprising four males and one female, aged between 23 and 29 years. Participants were informed that they would be shown a mix of AR and non-AR images and were tasked with providing responses based on the prompt T. To facilitate the study, we developed a custom Jupyter Notebook interface that displayed each image at 8 × 8 inches (20.32 × 20.32 cm), adjusted to fit the height of a laptop screen, allowing the entire image to be viewed clearly without scrolling. Participants' responses were recorded through the interface by clicking buttons to label their evaluation outcomes.

\noindent \textbf{Perception Performance:} 
The average True Positive Rate for Perception ($TPR_P$) across easy and medium levels closely aligned with the performance of VLMs, trailing the best $TPR_P$ of VLMs by only 1.7\% and 0.7\%, respectively. However, on hard-level examples, the average $TPR_P$ surpassed the best $TPR_P$ of VLMs by 8.1\%. This improvement can be attributed to the higher object density in hard-level images, which heightened participants’ suspicion and improved their ability to accurately identify virtual objects. For the same reason, each participant misclassified 1 to 8 non-AR images as AR images.

\noindent \textbf{Description Performance:} 
The average True Positive Rate for Description ($TPR_D$) across medium and hard levels exceeded the best $TPR_D$ of VLMs by 7.7\% and 27.1\%, respectively, while falling only 1.1\% short at the easy level. The superior performance at medium and hard levels can likely be attributed to humans' ability to learn from observed images and apply reasoning skills. Despite their enhanced reasoning, humans were prone to fatigue after reviewing approximately 100 images. Furthermore, we observed that the overall decline in human performance with increasing complexity levels closely mirrored the pattern exhibited by VLMs.

\noindent \textbf{Factors Influencing Performance:}
Recognition of AR content among participants was influenced by various factors, such as object texture, size, lighting, shadows, and positioning. Across the study, participants misclassified virtual objects as real or vice versa in 7.6\% to 13.8\% of the images, amounting to 24 to 44 misclassifications per participant. Interestingly, \emph{most misclassified images overlapped} with those where VLMs also struggled, indicating similarities in the challenges faced by humans and VLMs. Moreover, 2 participants showed a tendency to focus on objects that were closer and centrally positioned in the image, particularly during the later stages of the study. This finding provides guidance for fine-tuning VLMs to prioritize closer and central areas of images, aligning their attention more closely with human perception.

The study highlights both similarities and differences between human performance and VLMs, demonstrating the potential of VLMs to assess AR experience quality while also revealing areas where they fall short. To bridge this gap and align VLM performance more closely with human users, we plan to fine-tune VLMs using a manually curated dataset featuring diverse AR scenarios accompanied by corresponding textual descriptions.
\section{Conclusion and Future Work}
\label{sec:discussion} 
In this work, we evaluate VLMs for AR scene understanding using the DiverseAR dataset, the first AR dataset to assess VLMs' ability to identify and describe virtual content. Our results reveal that while VLMs perform well in detecting prominent AR elements, they struggle with seamlessly integrated virtual content, especially under conditions of realistic lighting, shadows, and physical plausibility.


Building on this study, we are enhancing the DiverseAR dataset by gathering a larger collection of AR images, each annotated with quantifiable properties such as virtual object size, contrast, and position, while also collecting human perception data for comparison. To further refine VLM performance, we are exploring AR-specific fine-tuning methods to align model capabilities with human perception, focusing on factors like shadow realism, lighting consistency, and object alignment. 

Beyond image-level assessment, we aim to expand our evaluation to include quality factors associated with temporal data modalities, such as the drift and jitter of virtual content. Building on the success of our previous non-VLM-based approaches~\cite{hu2024seesys, scargill2021will}, we will explore VLM-based methods to further analyze and assess these factors.

\acknowledgments{
We thank our user study participants for their invaluable assistance in this research. 
This work was supported in part by NSF grants CSR-2312760, CNS-2112562 and IIS-2231975, NSF CAREER Award IIS-2046072, NSF NAIAD Award 2332744, a CISCO Research Award, a Meta Research Award, Defense Advanced Research Projects Agency Young Faculty Award HR0011-24-1-0001, and the Army Research Laboratory under Cooperative Agreement Number W911NF-23-2-0224. The views and conclusions contained in this document are those of the authors and should not be interpreted as representing the official policies, either expressed or implied, of the Defense Advanced Research Projects Agency, the Army Research Laboratory, or the U.S. Government. This paper has been approved for public release; distribution is unlimited. No official endorsement should be inferred. The U.S. Government is authorized to reproduce and distribute reprints for Government purposes notwithstanding any copyright notation herein.}

\vspace{-0.2cm}

\bibliographystyle{abbrv-doi} 


\end{document}